\documentclass{article}
\pdfoutput=1 
\usepackage{nips13submit_e,times}
\nipsfinalcopy
\usepackage{hyperref}
\usepackage{url}
\usepackage[english]{babel}
\usepackage[utf8]{inputenc}
\usepackage{amsmath,amssymb}
\usepackage{graphicx}
\usepackage[colorinlistoftodos]{todonotes}
\usepackage{algorithm}
\usepackage{algpseudocode}
\usepackage{multirow}
\title{Deep Sequential Neural Network}

\author{Ludovic Denoyer and Patrick Gallinari}

\author{
Ludovic~Denoyer \\
Sorbonne Universites, UPMC Univ Paris 06, UMR 7606, LIP6, F-75005, Paris, France\\
			CNRS, UMR 7606, LIP6, F-75005, Paris, France \\
\texttt{ludovic.denoyer@lip6.fr} \\
\And
Patrick~Gallinari \\
Sorbonne Universites, UPMC Univ Paris 06, UMR 7606, LIP6, F-75005, Paris, France\\
			CNRS, UMR 7606, LIP6, F-75005, Paris, France \\
\texttt{patrick.gallinari@lip6.fr} \\
}
\date{\today}

\begin{document}
\maketitle

\begin{abstract}
Neural Networks sequentially build high-level features through their successive layers. We propose here a new neural network model where each layer is associated with a set of candidate mappings. When an input is processed, at each layer, one mapping among these candidates is selected  according to a sequential decision process. The resulting model is structured according to a DAG like architecture, so that a path from the root to a leaf node defines a sequence of transformations. Instead of considering global transformations, like in classical multilayer networks, this model allows us for learning a set of local transformations. It is thus able to process data with different characteristics through specific sequences of such local transformations, increasing the expression power of this model w.r.t a classical multilayered network. The learning algorithm is inspired from policy gradient techniques coming from the reinforcement learning domain and is used here instead of the classical back-propagation based gradient descent techniques. Experiments on different datasets show the relevance of this approach.

\end{abstract}

\section{Introduction}

Reinforcement Learning (RL) techniques which are usually devoted to problems in dynamic environments have been recently used for classical machine learning tasks like classification \cite{dulac2014sequentially,kegl12}. In that case, the prediction process is seen as a sequential process, and this sequential process can take different forms. For example \cite{dulac11} and \cite{graves14} consider that the sequential process is an acquisition process able to focus on relevant parts of the input data; \cite{darrell12} for example focuses on the sequential prediction process with a cascade approach. RL opens now some interesting research directions for classical ML tasks and allows one to imagine solutions to complex problems like budgeted classification \cite{dulac12} or anytime prediction \cite{think09}. 

In parallel, Neural Networks (NNs) have recently given rise to a large amount of research motivated by the development of deep architectures - or Deep Neural Networks (DNNs). The use of deep architectures have shown impressive results for many different tasks, from image classification \cite{deep1,deep2}, speech recognition \cite{deep3} to machine translation \cite{deep4} or even for natural language processing \cite{deep5}. These great successes mainly come from the ability of DNNs to compute high-level features over data. Many variants of learning algorithms have been proposed, from complex gradient computations \cite{algodeep1}, to dropout  methods \cite{algodeep2}, but the baseline learning algorithm still consists in recursively computing the gradient by using the back-propagation algorithm and performing (stochastic) gradient descent. 

This paper is motivated by the idea of using sequential learning algorithms - mainly coming from the reinforcement learning community - in the context of Deep Neural Networks. More precisely, we consider that inference in a NN is a sequential decision process which selects at each layer of a deep architecture one mapping among a set of candidate mappings. This process is repeated layerwise until the final layer is reached. The resulting NN  is then a DAG like architecture, where each layer is composed of a set of candidate mappings. Only one of these candidates will be selected at each layer, for processing an input pattern. When an input is presented to the NN, it will then follow a set of successive transformations which corresponds to a trajectory in the NN DAG, until the final output is computed. The decision on which trajectory to follow is computed at each layer through additional components called here selection functions. The latter are trained using a policy gradient technique like algorithm while the NN weights are trained using back propagation. This model called \textit{Deep Sequential Neural Networks} (DNNs) process an input through successive local transformations insetad of using a global transformation in a classical deep NN architecture. It can be considered as an extension of the classical deep NN architecture since when the number of potential candidate mapping at each layer is reduced to 1, one recovers a classical NN architecture. DSNNs are thus based on the following inference process:
\begin{itemize}
\item Given an input $x$, the model chooses between different possible mappings\footnote{We call a mapping the base transformation made between two layers of a neural network i.e a projection from $\mathbb{R}^n$ to $\mathbb{R}^m$.}
\item Then $x$ is mapped to a new representation space.
\item Given the new representation, another mapping is chosen between a set of different possible mappings, and so on to the prediction.
\end{itemize}
Note that the way mappings are chosen, and the mappings themselves are learned together on a training set. Instead of just computing representations in successive representation spaces, DSNNs are able to choose the best representation spaces depending on the input and to process differently data coming from different distributions.

This idea of choosing a different sequence of computations depending on the input share many common points with existing models (see Section \ref{sec:rw}) but our model has some interesting properties:
\begin{itemize}
\item It is able to simultaneously learn successive representations of an input, and also which representations spaces are the most relevant for this particular input.
\item Learning is made by extending \textbf{policy gradient} methods which as far as we know have never been used in this context; moreover, we show that, when the DNNs is in its simplest shape, this algorithm is equivalent to the gradient descent technique used in NNs.
\end{itemize}

The paper is organized as follows: in Section \ref{seq:model}, we describe the DSNN formalisms and the underlying sequential inference process. By deriving a policy gradient algorithm, we propose a learning algorithm in Section \ref{sec:learning} based on gradient descent techniques. We present in Section \ref{sec:exp} experimental results on different datasets and a qualitative study showing the ability of the model to solve complex classification problems. The related work is presented in Section \ref{sec:rw}.

\section{Deep Sequential Neural Networks}

\label{seq:model}
Let us consider $\mathcal{X} = \mathbb{R}^X$ the input space, and $\mathcal{Y}=\mathbb{R}^Y$ the output space, $X$ and $Y$ being respectively the dimension of the input and output spaces. We denote $\{(x_1,y_1),...,(x_\ell,y_\ell)\}$ the set of labeled training instances such that $x_i \in \mathcal{X} $ and $y_i \in \mathcal{Y}$. $\{(x_{\ell+1},y_{\ell+1}),...,(x_T,y_T)\}$ will denote the set of testing examples.\linebreak

The DSNN model has a DAG-structure defined as follow:
\begin{itemize}
\item Each node $n$ is in $\{n_1,...,n_N\}$ where $N$ is the total number of nodes of the DAG
\item The root node is $n_1$, $n_1$ does not have any parent node.
\item $c_{n,i}$ corresponds to the $i$-th child of node $n$ and $\#n$ is the number of children of $n$ so, in that case, $i$ is a value be between $1$ and $\#n$. 
\item $leaf(n)$ is \textit{true} if node $n$ is a leaf of the DAG - i.e a node without children.
\item Each node is associated to a particular representation space $\mathbb{R}^{dim(n)}$ where $dim(n)$ is the dimension associated to this space. Nodes play the same role than layers in classical neural networks.
\begin{itemize}
\item $dim(n_1) = X$ i.e the dimension of the root node is the dimension of the input of the model.
\item For any node $n$, $dim(n) = Y$ if $leaf(n)=true$ i.e the dimension of the leaf nodes is the output space dimension.
\end{itemize}
\item We consider \textit{mapping functions} $f_{n,n'} : \mathbb{R}^{dim(n)} \rightarrow \mathbb{R}^{dim(n')}$ which are functions associated with edge $(n,n')$. $f_{n,n'}$ computes a new representation of the input $x$ in node $n'$ given the representation of $x$ in node $n$. The output produced by the model is a sequence of $f$-transformation applied to the input like in a neural network.
\item In addition, each node is also associated with a \textit{selection function} denoted $p_n : \mathbb{R}^{dim(n)} \rightarrow \mathbb{R}^{\#n}$ able, given an input in $\mathbb{R}^{dim(n)}$, to compute a score for each child of node $n$. This function defines a probability distribution over the children nodes of $n$ such as, given a vector $z \in \mathbb{R}^{dim(n)}$
\begin{equation*}
P(c_{n,i} | z) = \frac{e^{p^i_n(z)}}{\sum\limits_{j=1}^{\#n} e^{p^j_n(z)}}
\end{equation*}
Selection functions aim at selecting which $f$-functions to use by choosing a path in the DAG from the root node to a leaf node.
\end{itemize}

\begin{figure}[t]
\begin{center}
\includegraphics[width=0.8\linewidth]{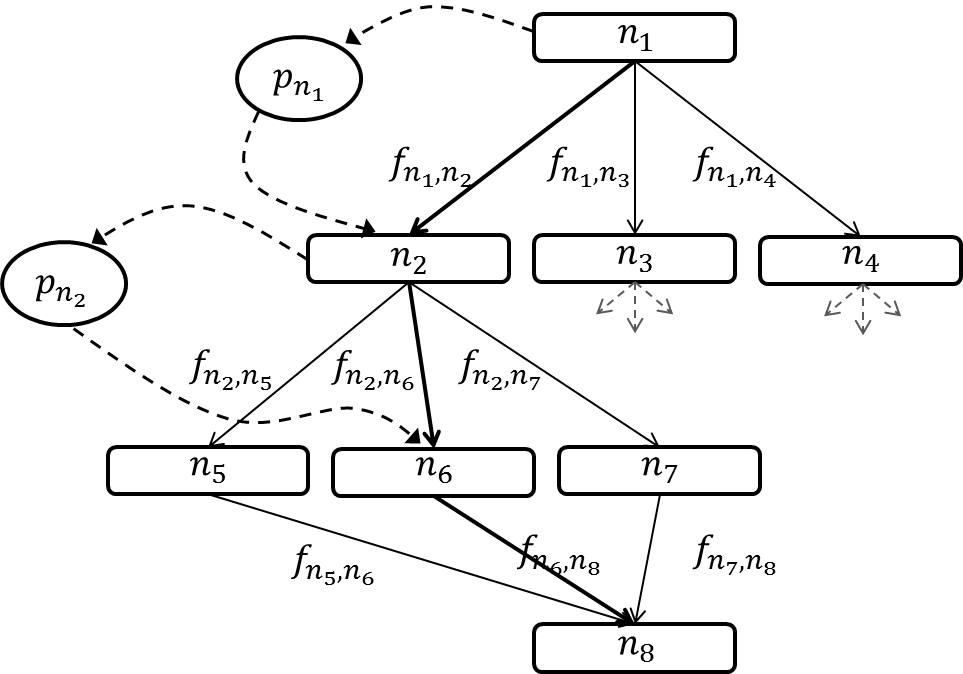}
\end{center}
\caption{Architecture of \textit{Deep Sequential Neural Networks}. We illustrate a model where each node has 3 children. For a particular input, and by using $p_{n_1}$ and $p_{n_2}$, the sequence of chosen nodes is $(n_1,n_2,n_6,n_8)$. Note that only $p_{n_1}$ and $p_{n_2}$ have been illustrated on the figure but each node is associated with a $p$ function. In that case, the final prediction given $x$ is $f_{n_6,n_8}(f_{n_2,n_6}(f_{n_1,n_2}(x)))$.}
\end{figure}

\subsection{Inference in DSNN}
Given such a DAG structure $\mathcal{G}$, the inference process is the following:
\begin{enumerate}
\item At first, an input $x \in \mathcal{X}$ is presented at the root node $n^{(1)}=n_1$ of the DAG\footnote{$n^{(t)}$ is used to denote the node selected at time $t$}.
\item Then, based on $x$, a child node $n^{(2)}$ is sampled using the $P(c_{^{(1)},.}|x)$ distribution computed through the $p_{n_1}$ function.
\item The model computes a new representation at node $n^{(2)}$ using $f_{n^{(1)},n^{(2)}}(x)$. A child node of $n^{(2)}$ is sampled following $P(c_{^{(2)},.}|x)$, .....
\item The same process is repeated until a leaf node. The vector computed at the leaf node level is the output of the model.
\end{enumerate}

Details of the inference procedure are given in Algorithm \ref{alg:inf}. The algorithm is a discrete-time sequential process starting at time $t=1$ and finishing when the input has reached a leaf node. Given an input $x$, we denote:
\begin{itemize}
\item $n^{(t)}$ the node reached by the input $x$ at time $t$ such that $n^{(1)}=n_1$.
\item $a^{(t)}$ the child node chosen at time $t$, $a^{(t)} \in [1..\#n^{(t)}]$
\item $z^{(t)}$ the mapping of $x$ at time $t$ such that $z^{(t)} \in \mathbb{R}^{dim(n^{(t)})}$
\end{itemize}

The inference process generates a trajectory $T$  which is a sequence $(n^{(1)},...,n^{(D)})$ of nodes starting from the root node $n^{(1)}=n_1$ to a leaf of the node $n^{(D)}$ such that $leaf(n^{(D)})=True$; $D$ is the size of the chosen branch of the tree. This sequence is obtained by sequentially choosing a sequence of children (or actions) $H=(a^{(1)},...,a^{(D-1)})$. In the following $H$ will denote a sequence of actions sampled w.r.t the $p$ functions.

%
%
%
\begin{algorithm}[t]
\caption{DSNN Inference Procedure}\label{alg_inference}
\begin{algorithmic}[1]
\Procedure{Inference}{$x$}\Comment{x is the input vector}
\State $z^{(1)} \gets x$
\State $n^{(1)} \gets n_1$
\State $t \gets 1$
\While{not $leaf(n^{(t)})$}\Comment{Inference finished}
\State $a^{(t)} \sim p_{n^{(t)}}(z^{(t)})$\Comment{Sampling using the distribution over children nodes}
\State $n^{(t+1)} \gets c_{n^{(t)},a^{(t)}}$
\State $z^{(t+1)} \gets f_{n^{(t)},n^{(t+1)}}(z^{(t)})$
\State $t  \gets t+1$
\EndWhile\label{euclidendwhile}
\State \textbf{return} $z^{(t)}$
\EndProcedure
\end{algorithmic}
\label{alg:inf}
\end{algorithm}

\section{Learning DSNN with gradient-based approaches}
\label{sec:learning}

The training procedure we propose aims at simultaneously learning both the \textit{mapping functions} $f_{i,j}$ and the \textit{selection functions} $p_i$ in order to minimize a given learning loss denoted $\Delta$. Our learning algorithm is based on an extension of \textbf{policy gradient techniques} inspired from the Reinforcement Learning literature. More precisely, our learning method is close to the methods proposed in \cite{schmi} and \cite{graves14} with the difference that, instead of considering a reward signal which is usual in reinforcement learning, we consider a loss function $\Delta$ computing the quality of the system. \hfill~\linebreak

Let us denote $\theta$ the parameters of the $f$ functions and $\gamma$ the parameters of the $p$ functions.

The performance of our system is denoted $J(\theta,\gamma)$:
\begin{equation}
J(\theta, \gamma) = E_{P(x,H,y)}[\Delta(F(x,H),y)]
\end{equation}
where both $H$ - i.e the sequence of children nodes chosen by the $p$-functions - and $F$ - the final decision given a particular path in the DSNN - depends on both $\gamma$ and $\theta$. The optimization of $J$ can be made by gradient-descent techniques and we need to compute the gradient of $J$: 
\begin{equation}
\begin{aligned}
\nabla_{\theta,\gamma} J(\theta,\gamma) &= \int \nabla_{\theta,\gamma} \left( P(H|x) \Delta(F(x,H),y) \right) P(x,y) dH dx dy \\
\end{aligned}
\end{equation}

This gradient can be simplified such that:
\begin{equation}
\begin{aligned}
\nabla_{\theta,\gamma} J(\theta,\gamma)& = \int \nabla_{\theta,\gamma} \left( P(H|x) \right) \Delta(F(x,H),y) P(x,y) dH dx dy + \int  P(H|x)  \nabla_{\theta,\gamma} \Delta(F(x,H),y) P(x,y) dH dx dy\\
&=  \int \frac{P(H|x)}{P(H|x)} \nabla_{\theta,\gamma} \left( P(H|x) \right) \Delta(F(x,H),y) P(x,y) dH dx dy \\
&+ \int  P(H|x)  \nabla_{\theta,\gamma} \Delta(F(x,H),y) P(x,y) dH dx dy\\
&=  \int P(H|x) \nabla_{\theta,\gamma} \left( log P(H|x) \right) \Delta(F(x,H),y) P(x,y) dH dx dy \\
&+ \int  P(H|x)  \nabla_{\theta,\gamma} \Delta(F(x,H),y) P(x,y) dH dx dy\\
\end{aligned}
\label{eq:eq}
\end{equation}

Using the Monte Carlo approximation of this expectation by taking $M$ trail histories over the training examples, we can write:
\begin{equation}
\nabla_{\theta,\gamma} J(\theta,\gamma) = \frac{1}{\ell}\sum\limits_{i=1}^{\ell} \left[  \frac{1}{M} \sum\limits_{k=1}^{M} \nabla_{\theta,\gamma} \left( log P(H | x_i) \right) \Delta(F(x_i,H),y) +  \nabla_{\theta,\gamma} \Delta(F(x_i,H),y) \right]
\end{equation}

Intuitively, the gradient is composed of two terms:
\begin{itemize}
\item The first term aims at penalizing trajectories with high loss - and thus encouraging to find trajectories with low loss. When the loss is $0$, the resulting gradient is null and the system will thus continue to choose the same paths.
\item The second term is the gradient computed over the branch of the tree that has been sampled. It encourages the $f$ functions to perform better the next time the same path will be chosen for a particular input.
\end{itemize}

While the second term can be easily computed by back-propagation techniques over the sequence of $f$ functions that compose the branch of the tree, the computation of $\nabla_{\theta,\gamma} log P(H | x_i)$ can be written:
\begin{equation}
\nabla_{\theta,\gamma}log P(H | x_i) = \nabla_{\theta,\gamma} \sum\limits_{t=1}^D log P(a^{(t)}|z^{(t)})
\end{equation}
The term $\nabla_{\theta,\gamma} log P(a^{(t)}|z^{(t)})$ depends on $z^{(t)}$ which is the projection of the input $x$ at node $n^{(t)}$. This projection involves the sequence of transformation $f_{n^{(1)},n^{(2)}}, ..., f_{n^{(t-1)},n^{(t)}}$ and the selection function $p_{n^{(t)}}$. It can also be computed by back-propagation techniques over the functions $f_{n^{(1)},n^{(2)}}, ..., f_{n^{(t-1)},n^{(t)}}, p_{n^{(t)}}$.

\paragraph{Variance reduction: } Note that equation \label{eq:eq} provides us an estimate of the gradient which can have a high variance. Instead of using this estimate,  
we replace $\Delta(F(x_i,H),y)$ by $\Delta(F(x_i,H),y)-b$ where $b = E_{p(x,H,y)}[\Delta(F(x_i,H),y)]$ which can be easily estimated on the training set \cite{schmi}.

\begin{figure}[t]
\begin{center}
\begin{tabular}{cc}
\includegraphics[width=0.5\linewidth]{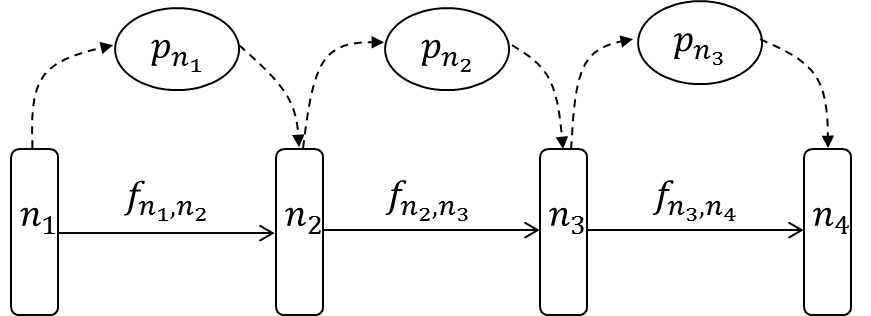} & \includegraphics[width=0.5\linewidth]{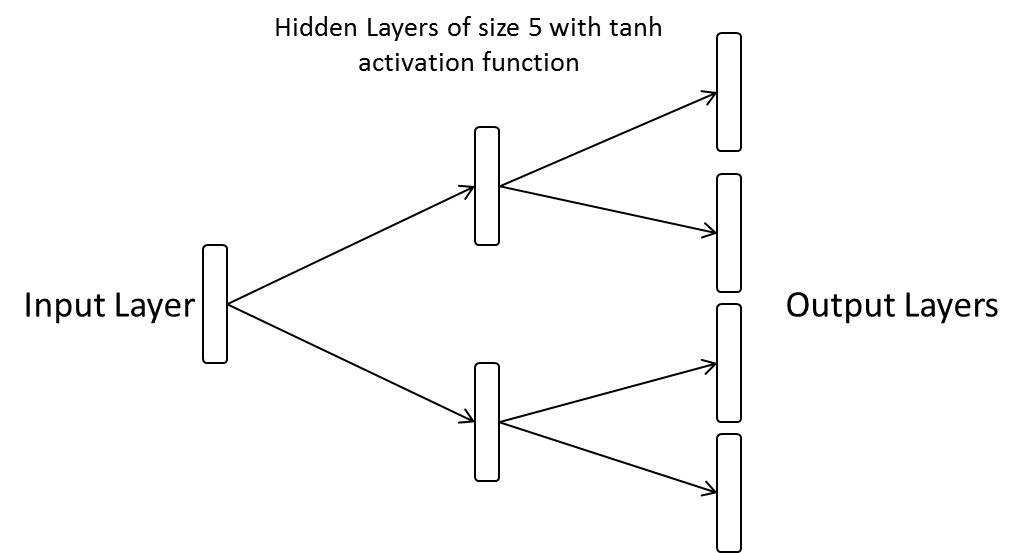} \\
(left) & (right)
\end{tabular}
\end{center}
\caption{(left) Architecture of the \textit{Deep Sequential Neural Network} with only one child for each node. In that case, the model is equivalent to a neural network. (right) Architecture of a \textit{DSNN-2 5 (tanh)} model}
\label{fig:nn1}
\end{figure}

\paragraph{NNs and DSNNs: } It is easy to show that DSNN is a generalization of NN and is equivalent to NN in its simple shape, where the structure is not a DAG but only a sequence of nodes as presented in Figure \ref{fig:nn1} (left). In other words, learning a DSNN with only one possible action at each timestep is equivalent to learning a neural network.

\begin{table}
\begin{center}
\begin{tabular}{cccc}
\multicolumn{4}{c}{No Hidden Layer (nhl)} \\
\includegraphics[width=0.2\linewidth]{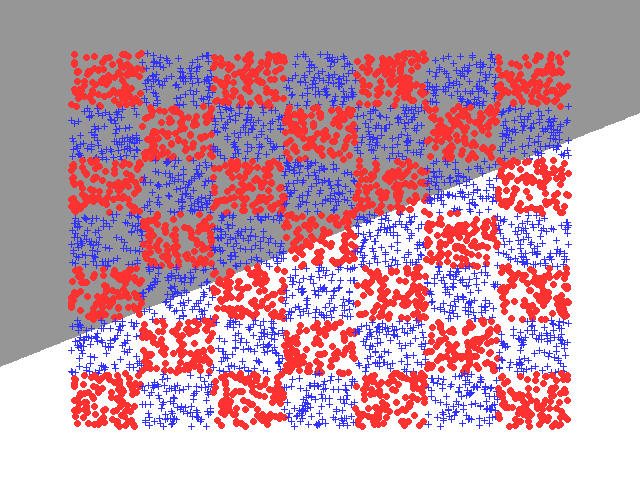} & \includegraphics[width=0.2\linewidth]{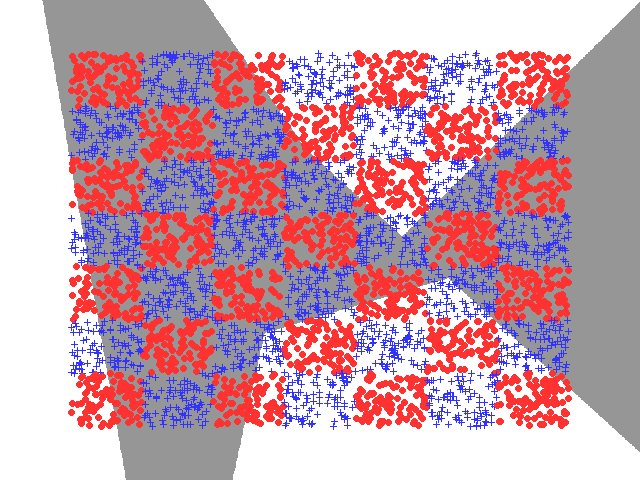} &\includegraphics[width=0.2\linewidth]{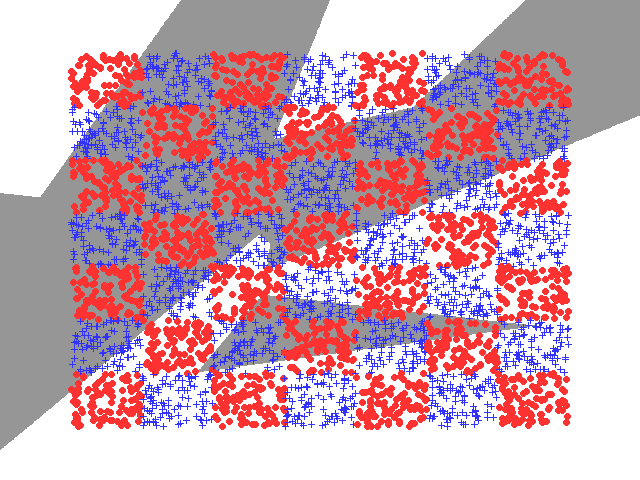} &\includegraphics[width=0.2\linewidth]{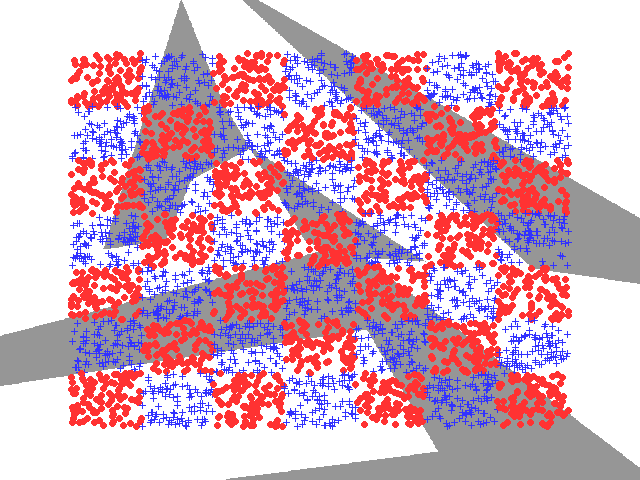}  \\
NN &  DSNN-3 &  DSNN-5 & DSNN-10 \\ 
\multicolumn{4}{c}{Hidden Layer Dimension: 10, Activation Function: Hyperbolic Tangent} \\
\includegraphics[width=0.2\linewidth]{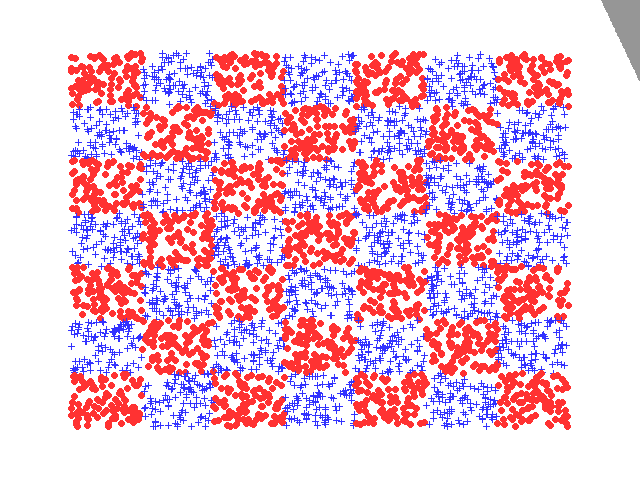} & \includegraphics[width=0.2\linewidth]{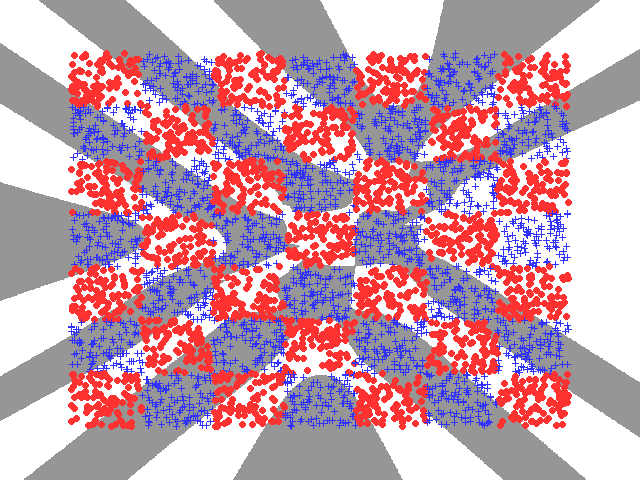} &\includegraphics[width=0.2\linewidth]{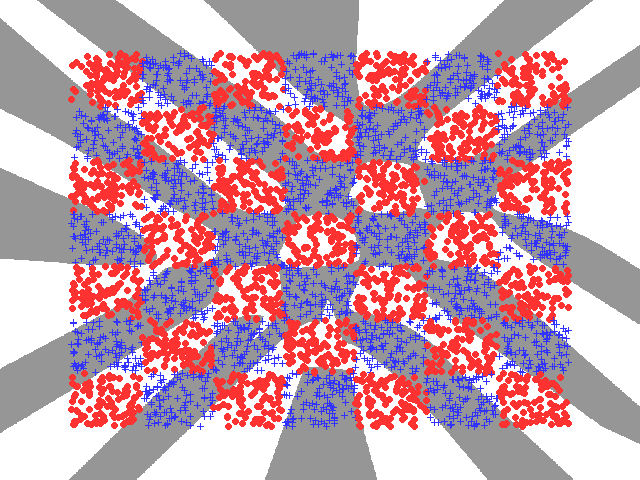} &\includegraphics[width=0.2\linewidth]{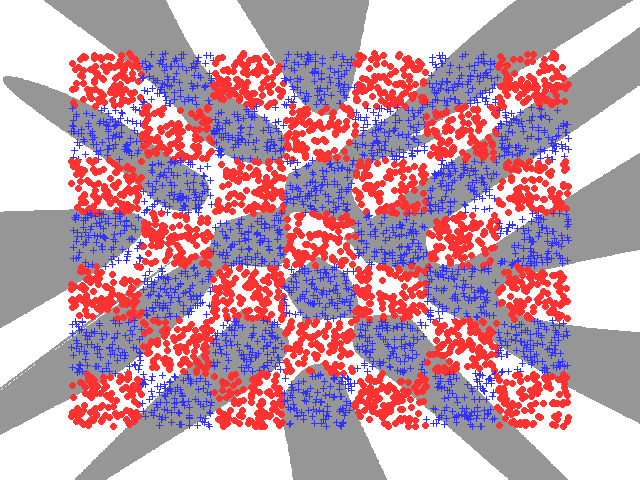}  \\
NN &   DSNN-3 &  DSNN-5 &  DSNN-10 \\ 
\multicolumn{4}{c}{Hidden Layers Dimension: 10-10, Activation Function: Hyperbolic Tangent} \\
\includegraphics[width=0.2\linewidth]{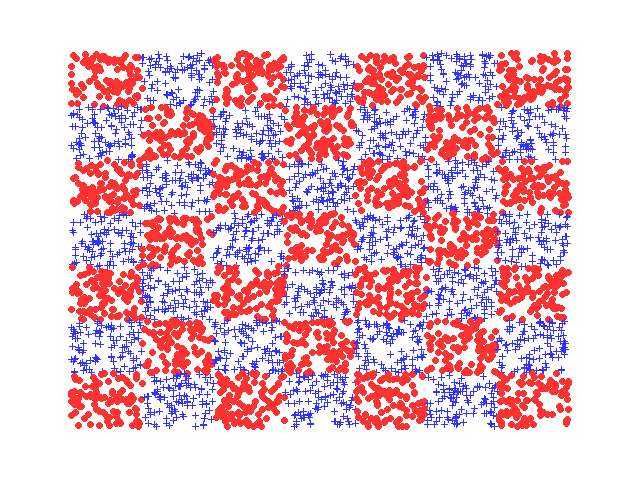} & \includegraphics[width=0.2\linewidth]{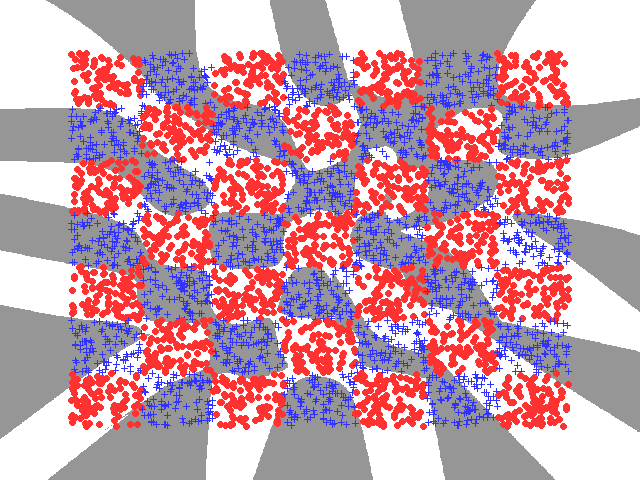} &\includegraphics[width=0.2\linewidth]{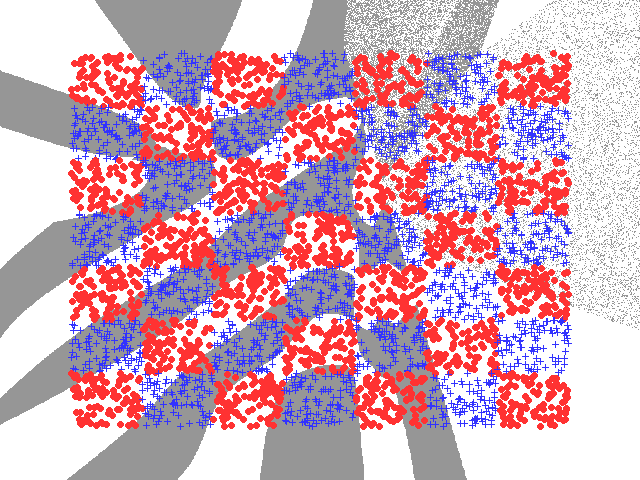} &\includegraphics[width=0.2\linewidth]{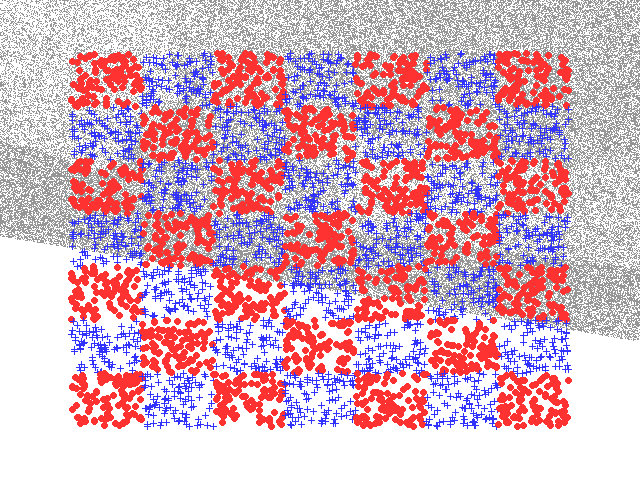}  \\
 NN &  DSNN-3 &  DSNN-5 &  DSNN-10 \\ 
\end{tabular}
\end{center}
\caption{Examples of Decision Frontiers obtained on the Checkboard $7 \times 7$ dataset }
\label{fig:ggrid}
\end{table}

\section{Experiment}
\label{sec:exp}

We have performed experiments comparing two different models: (i) \textbf{NN} corresponds to a simple neural network (ii) \textbf{DSNN-k} corresponds to the sequential model presented above where $k$ is the number of possible actions. The model corresponding to \textit{DSNN-2 5 (tanh)} is presented in Figure \ref{fig:nn1} (right). It corresponds to the extension of a NN with a 5-dimensionnal hidden layer (with hyperbolic tangent activation function) where now the system is able to choose at each timestep between 2 actions. $nhl$ will denote a model without hidden layer. \textit{DSNN-3 10-10 (rl)} corresponds to the extension of a NN with two hidden layers of size 10 (with rectified linear units) with 3 possible actions. The $f$ functions are thus linear transformations followed by a non linear function. The $p_.$ functions are simple linear functions\footnote{More complex $p$ functions could be used but have not been investigated in the paper}.

The experiments have been made on three families of datasets. The first set of experiments has been made on 5 UCI datasets which are datasets composed of about 1,000 training examples in low-dimensional space. The second set of experiments has been made on a variation of MNIST where the distribution of the inputs has been pertubated to measure the ability of the system to computes different features depending on the inputs. At last, the third set of experiments on simple 2-dimensionnal datasets based on checkerboard distributions and is used to better analyze the behavior of the model. The experiments have been performed with many different values of the hyper-parameters following a grid search algorithm. For each value of hyper-parameters and each dataset, we have performed 5 runs, we have averaged the performance over the 5 runs.

\begin{table}[h]
\begin{center}
\small{
        \begin{tabular}{rr||rrrrr}
          & \textbf{Hidden Layer(s)} & \multicolumn{1}{c}{\textbf{diabetes}} & \multicolumn{1}{c}{\textbf{fourclass}} & \multicolumn{1}{c}{\textbf{heart}} & \multicolumn{1}{c}{\textbf{sonar}} & \multicolumn{1}{c}{\textbf{splice}} \\ \hline \hline

    \multicolumn{1}{c}{\multirow{4}[2]{*}{\textbf{nhl}}} & \multicolumn{1}{l||}{NN} & \textbf{* 78.4} & 67.1 & \textbf{* 81.1} & \textbf{67.3} & 69.0 \\
    \multicolumn{1}{c}{} & \multicolumn{1}{l||}{DSNN-2} & 76.3 & 70.5 & 77.5 & 64.1 & \textbf{72.7} \\
    \multicolumn{1}{c}{} & \multicolumn{1}{l||}{DSNN-5} & 77.6 & 73.5 & 76.5 & 65.1 & 68.1 \\
    \multicolumn{1}{c}{} & \multicolumn{1}{l||}{DSNN-10} & 77.8 & \textbf{74.7} & 77.9 & 66.3 & 65.8 \\\hline
    \multicolumn{1}{c}{\multirow{4}[2]{*}{\textbf{5}}} & \multicolumn{1}{l||}{NN} & 75.8 & 76.5 & 69.6 & \textbf{79.7} & 60.0 \\
    \multicolumn{1}{c}{} & \multicolumn{1}{l||}{DSNN-2} & 75.2 & \textbf{94.5} & \textbf{74.3} & 77.1 & 68.7 \\
    \multicolumn{1}{c}{} & \multicolumn{1}{l||}{DSNN-5} & \textbf{76.9} & 92.1 & 72.8 & 79.4 & 70.5 \\
    \multicolumn{1}{c}{} & \multicolumn{1}{l||}{DSNN-10} & 76.5 & 92.7 & 70.3 & 77.5 & \textbf{69.8} \\\hline
    \multicolumn{1}{c}{\multirow{4}[2]{*}{\textbf{10}}} & \multicolumn{1}{l||}{NN} & \textbf{74.4} & 77.9 & 69.4 & 78.7 & 61.1 \\
    \multicolumn{1}{c}{} & \multicolumn{1}{l||}{DSNN-2} & 73.9 & \textbf{* 95.7} & 66.9 & 78.4 & 71.2 \\
    \multicolumn{1}{c}{} & \multicolumn{1}{l||}{DSNN-5} & 73.8 & 93.3 & 67.4 & \textbf{* 81.0} & \textbf{72.7} \\
    \multicolumn{1}{c}{} & \multicolumn{1}{l||}{DSNN-10} & 74.0 & 93.8 & \textbf{70.8} & 77.5 & 68.9 \\\hline
    \multicolumn{1}{c}{\multirow{4}[2]{*}{\textbf{25}}} & \multicolumn{1}{l||}{NN} & \textbf{77.1} & 77.0 & 68.1 & \textbf{77.5} & 61.3 \\
    \multicolumn{1}{c}{} & \multicolumn{1}{l||}{DSNN-2} & 73.3 & \textbf{94.4} & 72.8 & 77.1 & 67.8 \\
    \multicolumn{1}{c}{} & \multicolumn{1}{l||}{DSNN-5} & 72.6 & 93.1 & \textbf{73.3} & 75.9 & 69.1 \\
    \multicolumn{1}{c}{} & \multicolumn{1}{l||}{DSNN-10} & 73.4 & 91.0 & 70.6 & 76.8 & \textbf{71.4} \\\hline
    \multicolumn{1}{c}{\multirow{4}[2]{*}{\textbf{5-5}}} & \multicolumn{1}{l||}{NN} & 73.3 & 90.2 & 75.0 & \textbf{75.2} & 64.5 \\
    \multicolumn{1}{c}{} & \multicolumn{1}{l||}{DSNN-2} & 72.9 & \textbf{95.3} & 75.0 & 73.0 & \textbf{* 74.8} \\
    \multicolumn{1}{c}{} & \multicolumn{1}{l||}{DSNN-5} & 72.0 & 84.6 & \textbf{76.0} & 74.3 & 72.0 \\
    \multicolumn{1}{c}{} & \multicolumn{1}{l||}{DSNN-10} & \textbf{73.8} & 60.8 & 69.6 & 62.5 & 63.7 \\\hline
    \multicolumn{1}{c}{\multirow{4}[2]{*}{\textbf{10-10}}} & \multicolumn{1}{l||}{NN} & 70.1 & \textbf{92.8} & 77.2 & 76.8 & 64.6 \\
    \multicolumn{1}{c}{} & \multicolumn{1}{l||}{DSNN-2} & \textbf{73.6} & 90.0 & 73.5 & 75.2 & 71.0 \\
    \multicolumn{1}{c}{} & \multicolumn{1}{l||}{DSNN-5} & 72.0 & 88.7 & \textbf{77.7} & 76.2 & \textbf{73.5} \\
    \multicolumn{1}{c}{} & \multicolumn{1}{l||}{DSNN-10} & 73.0 & 67.4 & 76.7 & \textbf{79.4} & 67.4 \\\hline
    \multicolumn{1}{c}{\multirow{4}[2]{*}{\textbf{25-25}}} & \multicolumn{1}{l||}{NN} & 71.9 & \textbf{93.7} & 75.0 & 74.3 & 63.8 \\
    \multicolumn{1}{c}{} & \multicolumn{1}{l||}{DSNN-2} & 71.3 & 81.3 & 73.5 & \textbf{76.5} & \textbf{72.8} \\
    \multicolumn{1}{c}{} & \multicolumn{1}{l||}{DSNN-5} & 68.7 & 85.2 & \textbf{77.2} & 73.0 & 70.1 \\
    \multicolumn{1}{c}{} & \multicolumn{1}{l||}{DSNN-10} & \textbf{74.3} & 72.8 & 57.6 & 73.0 & 69.3 \\\hline
    \end{tabular}%
}
\end{center}
\caption{Accuracy over UCI datasets (with tanh activation function). $*$ is the best results obtained for each dataset. \textbf{Bold} values corresponds to the best performance obtained for each architecture. Results are average over 5 runs.}
\label{fig:uci}
\end{table}

\paragraph{UCI datasets: } The results obtained on UCI datasets are presented in Table \ref{fig:uci} where 50 \% of the examples have been used for training. First, one can see that, for some datasets (diabetes,heart), a simple linear model is sufficient for computing a high accuracy and using more complex architectures does not help in increasing the performance of the models. In that case, using DSNN does not seem really useful since a simple model is enough. For the other datasets, the DSNN outperforms the NN approach, particularly when the number of children for each node is low. Indeed, when this number becomes high, the number of parameters to learn can be very large, and the system is not able to learn these parameters, or needs at least much more iterations to converge.


\begin{table}
\begin{minipage}[b]{0.55\linewidth}
\small{
\vspace{-3cm}
    \begin{tabular}{r||r|r|r|r|}
          & \textbf{NN} & \textbf{DSNN-2} & \textbf{DSNN-3} & \textbf{DSNN-5} \\ \hline
    \multicolumn{1}{l||}{nhl} & 89.4 & 89.4 & 89.4 & 89.3 \\
    \multicolumn{1}{l||}{25} & 93.7 & 93.6 & 94.2 & 93.9 \\
    \multicolumn{1}{l||}{25-25} & 93.6 & 93.4 & 93.5 & 93.4 \\
    \multicolumn{1}{l||}{100} & 95.3 & 95.4 & 95.3 & 95.4 \\
    \multicolumn{1}{l||}{100-100} & 94.6 & 94.6 & 94.7 & 94.4 \\
    \end{tabular}%
}

\end{minipage}
\begin{minipage}[b]{0.5\linewidth}
\small{
        \begin{tabular}{r||r|r|r|r|}
    \textbf{} & \textbf{NN} & \textbf{DSNN-2} & \textbf{DSNN-3} & \textbf{DSNN-5} \\ \hline
    \multicolumn{1}{l||}{nhl} & 27.7 & 88.3 & 88.2 & \textbf{88.4} \\
    \multicolumn{1}{l||}{5} & 37.4 & 82.6 &  \textbf{83.5} & 56.7 \\
    \multicolumn{1}{l||}{10} & 83.4 &  \textbf{89.2} & 85.6 & 87.7 \\
    \multicolumn{1}{l||}{10-10} & 81.1 & \textbf{85.3} & 84.0 & 82.9 \\
    \multicolumn{1}{l||}{25} & \textbf{91.9} & 91.5 & 91.0 & 91.4 \\
    \multicolumn{1}{l||}{25-25} & \textbf{90.9} & 90.4 & 85.1 & 78.3 \\
		\multicolumn{1}{l||}{50-50} & 92.8 & \textbf{93.5} & 92.9 & 79.3 \\ 
    \end{tabular}%

}

\end{minipage}
\caption{Accuracy on the MNIST dataset (left) and the MNIST-Negative dataset (right) - digits have been resampled to $14 \times 14$ images. We have used rectified linear units on the hidden layers.}
\label{fig:mimi}
\end{table}

\paragraph{MNIST datasets: } We have performed experiments on both the classical MNIST dataset\footnote{The training set is composed of 12,000 examples, and the testing set is composed of 50,000 digits.} where digits have been re-sampled to $14 \times 14$ images, and to a variation of this dataset called MNIST-Negative where half of the digits have been negated - i.e for half of the digits, the value of a pixel is equal to one minus its original value. In that case, one can consider that digits have been sampled following two different distributions a simple model will not be able to capture. Table \ref{fig:mimi} reports the results we have obtained with different architectures. First, one can see that, for the MNIST dataset, the performance of NN and DSNN are quite similar showing that DSNN is not relevant when the input distribution is simple. On the MNIST-Inverse dataset, first, the NN without hidden layer is unable to well classify since the inputs are too much disparate. In that case, DSNN is able to capture the two inputs distributions and performs quite well. Adding some small hidden layers allows us to increase the accuracy. When using large hidden layers, a single NN is capable of capturing the data distribution and thus perform as well as DSNN. 

\paragraph{Checkerboard datasets: } For that set of experiments, we have generated checkerboard of points in two different categories (see Figure \ref{fig:ggrid}). The checkerboards sizes vary from $3 \times 3$ to $11 \times 11$ and each case of the checkerboard is composed of 100 training points, and 100 testing points. Performances are presented in Table \ref{fig:grid} showing that the DSNN model is able to capture this distribution. Figure \ref{fig:ggrid} show the decision frontiers obtained by different architectures. One can see that the NN model is not able to capture this distribution. DSNN-3 with a 10-dim hidden layer is almost perfect while DSNN models with a more complex architectures and a higher number of actions are not able to learn since they have too many parameters.

\begin{table}
\begin{center}
    \begin{tabular}{|r||r|r|r|r|r|r|r|r|r|r|} \hline
     Checkerboard: & \multicolumn{2}{c|}{ 3 $\times$ 3} & \multicolumn{2}{c|}{ 5 $\times$ 5} &   \multicolumn{2}{c|}{ 7$\times$ 7}  &    \multicolumn{2}{c|}{ 9$\times$9} &       \multicolumn{2}{c|}{ 11$\times$ 11} \\ \hline \hline
          & NN    & DSNN  & NN    &  DSNN & NN    &  DSNN & NN    &  DSNN & NN    &  DSNN \\ \hline
    Accuracy & 0.53  & 0.99     & 0.52  & 0.94  & 0.52  & 0.86  & 0.51  & 0.749 & 0.5   & 0.697 \\  \hline
    
    \end{tabular}%

\end{center}
\caption{Performance over the checkerboard datasets. Only the best performance have been reported. NNs and DSNNs have been tested with the following architectures: nhl,$2$,$5$,$10$,$2-2$,$5-5$,$10-10$. DSNNs have been tested with 2, 3, 5 and 10 possible actions.  }
\label{fig:grid}
\end{table}

\section{Related Work}
\label{sec:rw}

Different models are related to DSNNs. The first family of models are neural networks. The idea of processing input data by different functions is not new and have been proposed for example in Neural Tree Networks \cite{ptree,nt}, with Hierarchical Mixture of Experts \cite{Jordan} where the idea is to compute different transformations of data and to aggregate these transformations. The difference with our approach is both in the inference process, and in the way the model is learned. They also share the idea of processing different inputs with different computations which is the a major idea underlying decision trees \cite{dt} and also more recent classification techniques like \cite{dulac11}. 

At last, some links have already be done between classification and reinforcement learning algorithms \cite{dulac12,kegl12}. Particularly, the use of recurrent neural networks from modelling Markov Decision Processes learned by Policy gradient techniques has been deeply explored in \cite{schmi} and in a recent work that proposes the use of such models for image classification \cite{graves14}.

\section{Conclusion and Perspectives}

We have proposed a new family of model called \textit{Deep Sequential Neural Networks} which differ from neural networks since, instead of always applying the same set of transformations, they are able to choose which transformation to apply depending on the input. The learning algorithm is based on the computation of the gradient which is obtained by an extension of policy-gradient techniques. In its simplest shape, DSNNs are equivalent to DNNs. Experiments on different datasets have shown the effectiveness of these models.
\pagebreak
\bibliographystyle{plain}
\bibliography{bib}

\begin{thebibliography}{10}

\bibitem{algodeep2}
Pierre Baldi and Peter~J. Sadowski.
\newblock The dropout learning algorithm.
\newblock {\em Artif. Intell.}, 210:78--122, 2014.

\bibitem{kegl12}
R{\'{o}}bert Busa{-}Fekete, Djalel Benbouzid, and Bal{\'{a}}zs K{\'{e}}gl.
\newblock Fast classification using sparse decision dags.
\newblock In {\em Proceedings of the 29th International Conference on Machine
  Learning, {ICML} 2012, Edinburgh, Scotland, UK, June 26 - July 1, 2012},
  2012.

\bibitem{dulac11}
Gabriel Dulac{-}Arnold, Ludovic Denoyer, and Patrick Gallinari.
\newblock Text classification: {A} sequential reading approach.
\newblock In {\em Advances in Information Retrieval - 33rd European Conference
  on {IR} Research, {ECIR} 2011, Dublin, Ireland, April 18-21, 2011.
  Proceedings}, pages 411--423, 2011.

\bibitem{dulac12}
Gabriel Dulac{-}Arnold, Ludovic Denoyer, Philippe Preux, and Patrick Gallinari.
\newblock Sequential approaches for learning datum-wise sparse representations.
\newblock {\em Machine Learning}, 89(1-2):87--122, 2012.

\bibitem{dulac2014sequentially}
Gabriel Dulac-Arnold, Ludovic Denoyer, Nicolas Thome, Matthieu Cord, and
  Patrick Gallinari.
\newblock Sequentially generated instance-dependent image representations for
  classification.
\newblock {\em Internation Conference on Learning Representations - ICLR 2014},
  2014.

\bibitem{think09}
Alireza Farhangfar, Russell Greiner, and Csaba Szepesv{\'{a}}ri.
\newblock Learning to segment from a few well-selected training images.
\newblock In {\em Proceedings of the 26th Annual International Conference on
  Machine Learning, {ICML} 2009, Montreal, Quebec, Canada, June 14-18, 2009},
  page~39, 2009.

\bibitem{deep3}
Alex Graves, Abdel{-}rahman Mohamed, and Geoffrey~E. Hinton.
\newblock Speech recognition with deep recurrent neural networks.
\newblock In {\em {IEEE} International Conference on Acoustics, Speech and
  Signal Processing, {ICASSP} 2013, Vancouver, BC, Canada, May 26-31, 2013},
  pages 6645--6649, 2013.

\bibitem{Jordan}
Michael~I. Jordan and Robert~A. Jacobs.
\newblock Hierarchical mixtures of experts and the em algorithm.
\newblock {\em Neural Comput.}, 6(2):181--214, March 1994.

\bibitem{darrell12}
Sergey Karayev, Tobias Baumgartner, Mario Fritz, and Trevor Darrell.
\newblock Timely object recognition.
\newblock In {\em Advances in Neural Information Processing Systems 25}, pages
  899--907, 2012.

\bibitem{deep1}
Alex Krizhevsky, Ilya Sutskever, and Geoffrey~E. Hinton.
\newblock Imagenet classification with deep convolutional neural networks.
\newblock In {\em Advances in Neural Information Processing Systems 25}, pages
  1106--1114, 2012.

\bibitem{algodeep1}
James Martens and Ilya Sutskever.
\newblock Training deep and recurrent networks with hessian-free optimization.
\newblock In {\em Neural Networks: Tricks of the Trade - Second Edition}, pages
  479--535. 2012.

\bibitem{graves14}
Volodymyr Mnih, Nicolas Heess, Alex Graves, and Koray Kavukcuoglu.
\newblock Recurrent models of visual attention.
\newblock {\em CoRR}, abs/1406.6247, 2014.

\bibitem{dt}
J.~Ross Quinlan.
\newblock Induction of decision trees.
\newblock {\em Machine Learning}, 1(1):81--106, 1986.

\bibitem{deep2}
Jurgen Schmidhuber.
\newblock Multi-column deep neural networks for image classification.
\newblock In {\em Proceedings of the 2012 IEEE Conference on Computer Vision
  and Pattern Recognition (CVPR)}, CVPR '12, pages 3642--3649, Washington, DC,
  USA, 2012. IEEE Computer Society.

\bibitem{nt}
J~A Sirat and J-P Nadal.
\newblock Neural trees: a new tool for classification.
\newblock {\em Network: Computation in Neural Systems}, 1(4):423--438, 1990.

\bibitem{deep5}
Richard Socher and Christopher~D. Manning.
\newblock Deep learning for {NLP} (without magic).
\newblock In {\em Human Language Technologies}, pages 1--3, 2013.

\bibitem{ptree}
Paul~E. Utgoff.
\newblock Perceptron trees: {A} case study in hybrid concept representations.
\newblock In {\em Proceedings of the 7th National Conference on Artificial
  Intelligence. St. Paul, MN, August 21-26, 1988.}, pages 601--606, 1988.

\bibitem{schmi}
Daan Wierstra, Alexander F{\"{o}}rster, Jan Peters, and J{\"{u}}rgen
  Schmidhuber.
\newblock Solving deep memory pomdps with recurrent policy gradients.
\newblock In {\em Artificial Neural Networks - {ICANN} 2007}, pages 697--706,
  2007.

\bibitem{deep4}
Will~Y. Zou, Richard Socher, Daniel~M. Cer, and Christopher~D. Manning.
\newblock Bilingual word embeddings for phrase-based machine translation.
\newblock In {\em Proceedings of the 2013 Conference on Empirical Methods in
  Natural Language Processing, {EMNLP}}, pages 1393--1398, 2013.

\end{thebibliography}

\end{document}